\documentclass[conference]{IEEEtran}
\IEEEoverridecommandlockouts
\usepackage{cite}
\usepackage{amsmath,amssymb,amsfonts}
\usepackage{algorithmic}
\usepackage{graphicx}
\usepackage{booktabs}
\usepackage{makecell}
\usepackage{colortbl}  
\usepackage[hidelinks]{hyperref}
\usepackage{textcomp}
\usepackage{xcolor}
\def\BibTeX{{\rm B\kern-.05em{\sc i\kern-.025em b}\kern-.08em
    T\kern-.1667em\lower.7ex\hbox{E}\kern-.125emX}}
\begin{document}

\title{Automatic Labelling \& Semantic Segmentation with 4D Radar Tensors\\

}

\author{\IEEEauthorblockN{Botao Sun, Ignacio Roldan, Francesco Fioranelli}
\IEEEauthorblockA{\textit{Microwave Sensing, Signals \& Systems (MS3) Group, Dept. of Microelectronics, TU Delft} \\
Delft, The Netherlands \\
\IEEEauthorblockA{ \emph{email:B.Sun-6@student.tudelft.nl; i.roldanmontero@tudelft.nl; f.fioranelli@tudelft.nl}}
}
%\and
%\and
}

\maketitle

\begin{abstract}
In this paper, an automatic labelling process is presented for automotive datasets, leveraging on complementary information from LiDAR and camera. The generated labels are then used as ground truth with the corresponding 4D radar data as inputs to a proposed semantic segmentation network, to associate a class label to each spatial voxel. Promising results are shown by applying both approaches to the publicly shared \textit{RaDelft} dataset, with the proposed network achieving over 65\% of the LiDAR detection performance, improving 13.2\% in vehicle detection probability, and reducing 0.54 m in terms of Chamfer distance, compared to variants inspired from the literature.

\end{abstract}


\section{Introduction}
% \textbf{HERE GOES AN INTRO;HERE GOES AN INTRO;HERE GOES AN INTRO;HERE GOES AN INTRO;HERE GOES AN INTRO;HERE GOES AN INTRO;v;HERE GOES AN INTRO;HERE GOES AN INTRO;HERE GOES AN INTRO;HERE GOES AN INTRO;HERE GOES AN INTRO;HERE GOES AN INTRO;HERE GOES AN INTRO;HERE GOES AN INTRO;HERE GOES AN INTRO;HERE GOES AN INTRO;HERE GOES AN INTRO;HERE GOES AN INTRO;HERE GOES AN INTRO;HERE GOES AN INTRO;HERE GOES AN INTRO;HERE GOES AN INTRO;HERE GOES AN INTRO;HERE GOES AN INTRO;HERE GOES AN INTRO;HERE GOES AN INTRO;HERE GOES AN INTRO;HERE GOES AN INTRO;HERE GOES AN INTRO} 

Radar's robustness in adverse weather and ability to provide dynamic information makes it a valuable complement to cameras and LiDAR in advanced driver-assistance systems (ADAS) \cite{wang2019multi}. Although deep learning methods for semantic segmentation are well-developed for RGB images and LiDAR point clouds (PCs), their application in radar remains underexplored, especially with 4D radar data that includes additional elevation information \cite{chen2017rethinking}\cite{dosovitskiy2020image}\cite{qi2017pointnet++}\cite{zhao2021point}. This paper addresses this research gap by proposing a method to perform semantic segmentation directly on the range-azimuth-elevation-Doppler (RAED) tensor. Additionally, a novel automatic labelling process is introduced to generate point-by-point multi-class labels in the \textit{RaDelft} dataset, thus enabling joint detection and classification using radar data. %Additional details are presented in \cite{botao_thesis}.

\section{Proposed automatic labelling process}
%This section introduces the proposed automatic labelling process, illustrated in Fig.~\ref{fig: Label block diagram}, and presents results based on the \textit{RaDelft} dataset. 
%first introduces the RaDelft dataset and the need for the automatic labelling process. We then detail the method, covering preliminary object label generation, calibration, coordinate transformation, and voxelization. The whole process is illustrated in Fig.~\ref{fig: Label block diagram}. Finally, we compare the quantitative and visualization results of automatic versus manual labelling.
%\subsection{Brief introduction of the RaDelft dataset}
The work in this paper uses the \textit{RaDelft} dataset, with 4D radar tensors, LiDAR Point Clouds, and RGB images for 16975 frames across 7 different driving scenes recorded in Delft. Additional details on the dataset collection setup are discussed in \cite{roldan2024deepautomotiveradardetector}. 
4D radar tensors are generated from the raw data by applying two 2D FFTs to create RAED tensors. Each radar tensor has two channel dimensions: a Range-Azimuth-Doppler (RAD) tensor containing power values, and another one containing single elevation values from 1 to 34, corresponding to the highest power return. Consequently, the dimension of the RAED tensor is $2\times 128 \times 240 \times 500$ ($N_C \times N_D \times N_A \times N_R$), representing channel, Doppler, azimuth, and range, respectively.
%LiDAR and radar sensors are installed parallel on the car roof, with the camera positioned on the windshield; this aligns the perception areas of the radar and LiDAR, facilitating transformations through translation and rotation. All sensors have been spatially calibrated using reference targets, enabling projection, translation, and rotation of recorded data across sensors. 

\begin{figure*}[ht]
\centering
\includegraphics[width=0.8\textwidth]{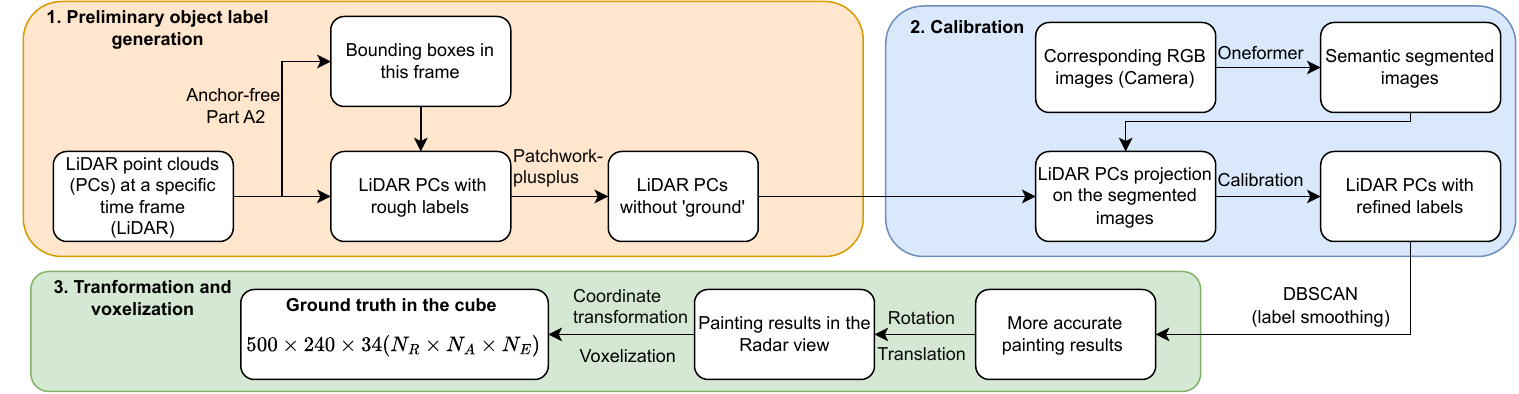}
\caption{Block diagram of the proposed automatic labelling process using LiDAR point clouds (PCs) and RGB images \cite{botao_thesis}. First, preliminary labels on LiDAR PCs are generated by a pre-trained object detection model. Semantic information from the images is then used to calibrate key labels in the central view, followed by label consistency adjustment using DBSCAN. Finally, point-by-point multi-class labels are generated by coordinate transformation and voxelization.}
\label{fig: Label block diagram}

\end{figure*}

\begin{figure*}[ht]
\centering
\includegraphics[width=0.7\textwidth]{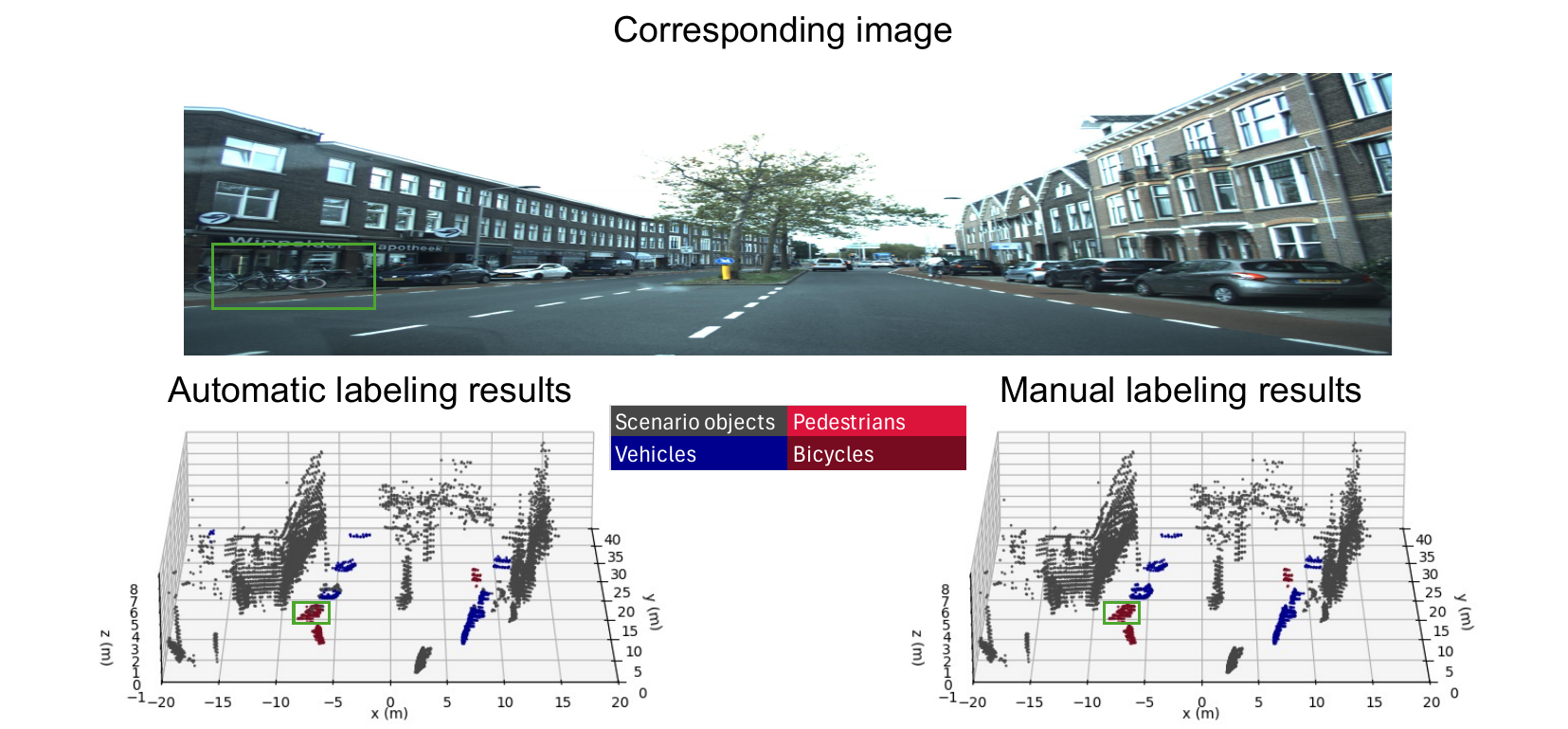}
\caption{Automatic vs manual labelling results for a complex scene, with a reference camera image provided. The color bar distinguishes the 4 labelled classes.}
%The automatic labelling closely aligns with the manual results, with minor discrepancies observed in the green boxes where closely placed bicycles lead to slight variations.}

\label{fig: Visualization results of labelling process}
\end{figure*}

For semantic segmentation, point-by-point multi-class labels are necessary. However, the \textit{RaDelft} dataset does not provide labels, hence an automatic labelling process is proposed and explained in the following sections. The whole process is illustrated in Fig.~\ref{fig: Label block diagram}. %First, LiDAR PCs are translated and rotated to align with the radar view, serving as ground truth. RGB images are employed for calibration by projecting the LiDAR PCs onto them. 
%was originally created for a data-driven detector to generate LiDAR-like PCs from radar cubes, using LiDAR PCs as ground truth without class labels.  To address this, an automatic labelling process is developed to significantly reduce labelling costs. 

\subsection{Automatic labelling approach}
\subsubsection{Preliminary object label generation}

Benefiting from the rapid development of 3D object detection only using LiDAR PCs, rough object labels can be generated by applying a pre-trained detection model directly to the LiDAR PCs. Here, we use the anchor-free `Part-A2 model' \cite{shi2020points} trained on the KITTI dataset \cite{Geiger2013IJRR}. %The anchor-free method is suitable for dealing with irregular scenes. Compared to the anchor-based strategy, it is lighter and more memory-efficient. It also avoids the need to set bounding box dimensions before processing. This is particularly important for detecting small-scale pedestrians and bicycles at long distances from the LiDAR.
As the pre-trained model is not specifically designed or trained for the \textit{RaDelft} dataset, its performance is not optimized. The KITTI dataset uses a 64-layer Velodyne LiDAR, while the \textit{RaDelft} uses a 128-layer RoboSense LiDAR, resulting in denser PCs. This increased density particularly affects the detection of nearby bicycles, which may be misclassified as ground or buildings. To mitigate this, we first downsample the LiDAR PCs to reduce their density before applying the pre-trained model.  
A high-confidence threshold of 0.5 is set to ensure detection accuracy and retain only bounding boxes with class information. Using seven bounding box parameters such as center coordinates $x, y, z$, bounding box size $d_x, d_y, d_z$, and rotation angle $heading$ along the z-axis, the vertex coordinates in Cartesian coordinates are generated. Points within each bounding box are then assigned the corresponding class.
%However, this results in some short poles being misclassified as pedestrians. To address this issue, a higher confidence threshold of $0.8$ is applied for pedestrians beyond $30 m$ from the LiDAR. Although some distant pedestrians may be missed, radar also struggles to detect them effectively.

%Currently, the LiDAR PCs $L \in \mathbb{R}^{N,4+C}$, with $N$ points and $C$ classes, retain $360^\circ$ field of view (FoV) and $200$ meters detection range, with the forward targets labeled. 
To align with the radar parameters, the whole LiDAR PCs are then cropped to fit the radar view with an azimuth angle from $-70^\circ$ to $70^\circ$, a maximum unambiguous range of 51.4 m, and an elevation angle from $-15^\circ$ to $15^\circ$. In this cropping process, the transformation from LiDAR to radar view is applied.
%These two sensors are assembled in parallel to ensure the LiDAR view covers the radar view. 
After cropping, the PCs are transformed back to the LiDAR view for calibration.
%to use the extrinsic calibration parameters $\mathbf{T}^{\text{LiDAR},\text{Camera}}$ during calibration. 
Notably, LiDAR detects ground points, which can form dense clusters and negatively impact radar detection. To mitigate this, the ground points can be adaptively eliminated by `Patchwork++' \cite{lee2022patchworkpp} so that only four classes of targets remain, i.e., `scenario objects', `pedestrians', `vehicles', and `bicycles'.
%resulting in $L_\text{reduced} \in \mathbb{R}^{N'}$. After this, only the `scenario objects', `pedestrians', `vehicles', and `bicycles' are preserved for further processing.

\subsubsection{Calibration process}
Despite setting a high confidence threshold, some mis-labelling events still occur. To address this, RGB images can be used for calibration. As the camera is placed directly below the radar, this alignment ensures that its view overlaps with the central field of view of the radar, which is more crucial than the side view. We employ the transformer-based pre-trained model `OneFormer' \cite{jain2023oneformer} to extract semantic information from camera data. 
%From the RGB image $I\in\mathbb{R}^{W, H}$ where $W$ represents width and $H$ represents height, semantic information $S_\text{semantic} \in  \mathbb{R}^{W, H, C}$ is obtained where $C$ represents the number of classes. 
Inspired by \cite{domhof2021joint}, extrinsic calibration parameters between the camera and LiDAR can be used to estimate the relative rigid transformation between them, i.e., a projection from the 3D LiDAR PCs to the 2D camera images to align the PCs with the image pixels. %$\mathbf{T}^{\text{LiDAR},\text{Camera}} \in 4\times 4$, enabling the transformation from the LiDAR view to the camera view. 
A 25-m distance threshold is also set to use semantic information from the camera as labels, so that the new labels for the points close to the sensors are assigned to labels generated from the camera, while the distant points retain their original labels.
%L_\text{reduced}$, while the distant points retain their original labels.

Despite combining semantic information from two sensors and focusing on central view targets, labelling imperfections remain, such as incomplete labelling of large objects due to small bounding boxes. To address this, label consistency adjustment is applied using the Density-Based Spatial Clustering of Applications with Noise (DBSCAN) algorithm to identify clusters in the PCs \cite{ester1996density}. In each cluster, labels are assigned to the same class via majority voting. To detect small clusters and differentiate closely located targets, we set $\epsilon = 0.6$, and $minPts = 100$. Finally, refined labels are generated.

\subsubsection{Tranformation and voxelization}
As the radar data is structured as a cube in polar coordinates, directly using non-uniform LiDAR PCs in Cartesian coordinates as labels is infeasible. Thus, we first rotate and translate the LiDAR PCs to align with the radar view and voxelize the LiDAR PCs to convert them into a cube with the dimension $500 \times 240 \times 34$ ($N_R \times N_A \times N_E$) representing range, azimuth, \& elevation, respectively. Each voxel in the cube requires a label, defined by majority voting as 0 for \textbf{`empty'}, 1 for \textbf{`scenario objects'}, 2 for \textbf{`pedestrians'}, 3 for \textbf{`vehicles'}, and 4 for \textbf{`bicycles'}.

%During voxelization, majority voting determines the label for each voxel.  If a voxel is empty, it is assigned a label of 0. Labels are assigned as follows: $1$ for \textbf{`scenario objects'}, $2$ for \textbf{`pedestrians'}, $3$ for \textbf{`vehicles'}, and $4$ for \textbf{`bicycles'}. 
%In the voxelization process, the range axis is uniform, with each range cell size $0.1004 m$, while the non-uniform voxelization is applied in the azimuth and elevation axis. Consequently, the side bins in the azimuth and elevation axes are sparse, and the central bins are dense, leading to the preservation of more central targets compared to the side targets.

\begin{table}[htbp]
\centering
\caption{Quantitative metrics of the designed automatic labelling results compared to manual labelling results.}
\begin{tabular}{lccc}
\toprule
\textbf{Class} & \textbf{Precision} & \textbf{Recall} & \textbf{F1-score} \\ 
\midrule
Scenario objects      & 0.98 & 0.99  &0.99 \\ 
Vehicles  & 0.95 & 0.81  & 0.88 \\ 
Bicycles      & 0.79 & 0.87  & 0.83 \\ 
Pedestrians        & 0.67 & 0.72  & 0.69 \\ 
\bottomrule
\end{tabular}
\label{tab1}
\end{table}

\subsection{Automatic labelling results}
\subsubsection{Qualitative results}
Fig.~\ref{fig: Visualization results of labelling process} compares the proposed automatic vs manual labelling results for a sample frame. The majority of points are correctly labeled in the automatic labelling process, with the exception of some points within the green box highlighted in the figure. There, multiple bicycles are closely parked, complicating the generation of bounding boxes during the preliminary object labelling process. However, through the calibration process, points within the 25m range are correctly calibrated to the `bicycles' class.

\subsubsection{Quantitative results}
%Given the large dataset of nearly 17,000 frames, it is impractical to manually check each one. 
For a quantitative evaluation of the proposed automatic labelling, we randomly selected 50 frames from 7 different scenes for manual expert labelling. Precision, recall, and F1-score metrics were then used to compare the automatic labelling with the manual labels. The results are shown in Table \ref{tab1}. The performance for `scenario objects' and `vehicles' is satisfactory, while the performance for `bicycles' and `pedestrians' is slightly lower, which is expected with the greater difficulty in automatically labelling smaller targets.

\section{Proposed Semantic Segmentation Network}
%This section details the semantic segmentation network using 4D radar cubes. The process begins with the pre-processing of radar cubes to prepare them as network inputs. Next, the network architecture is presented, followed by a discussion of key training details. Finally, the network performance is evaluated and compared with other methods.

%\subsection{Data pre-processing}

\begin{figure}[htbp]
\centering
\includegraphics[width=0.7\columnwidth]{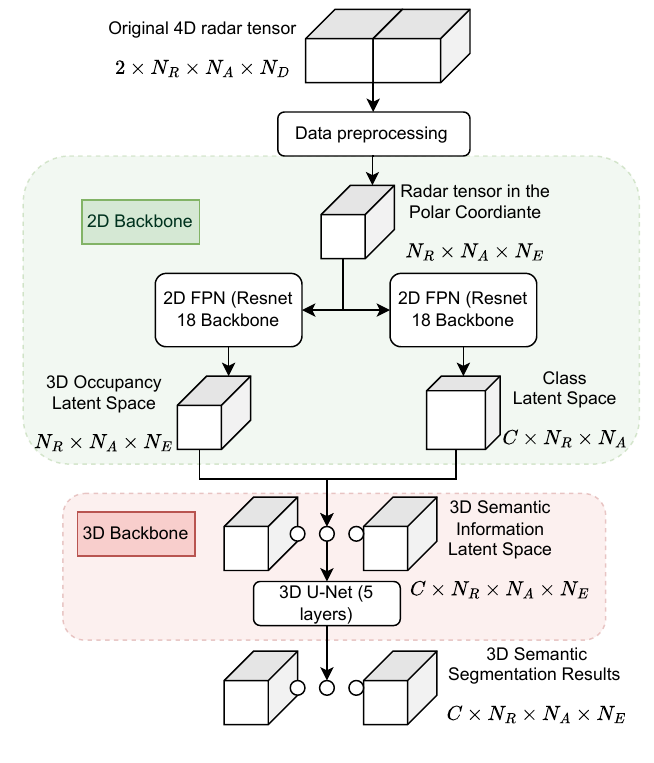}
\caption{Proposed radar semantic segmentation approach \cite{botao_thesis}. A radar tensor in polar coordinates with dimensions $N_R \times N_A \times N_E$ is generated after data preprocessing. Next, a 2D backbone with two individual branches is constructed to generate the 3D occupancy latent space and the class latent space, respectively. These latent spaces are then combined through broadcasting to form a new 3D semantic segmentation latent space. Finally, a 3D backbone is used to produce the output, providing the probability of occurrence for each class at every voxel.}

\label{fig: Block diagram of network structure}
\end{figure}

\begin{figure*}[ht]
\centering
\includegraphics[width=0.7\textwidth]{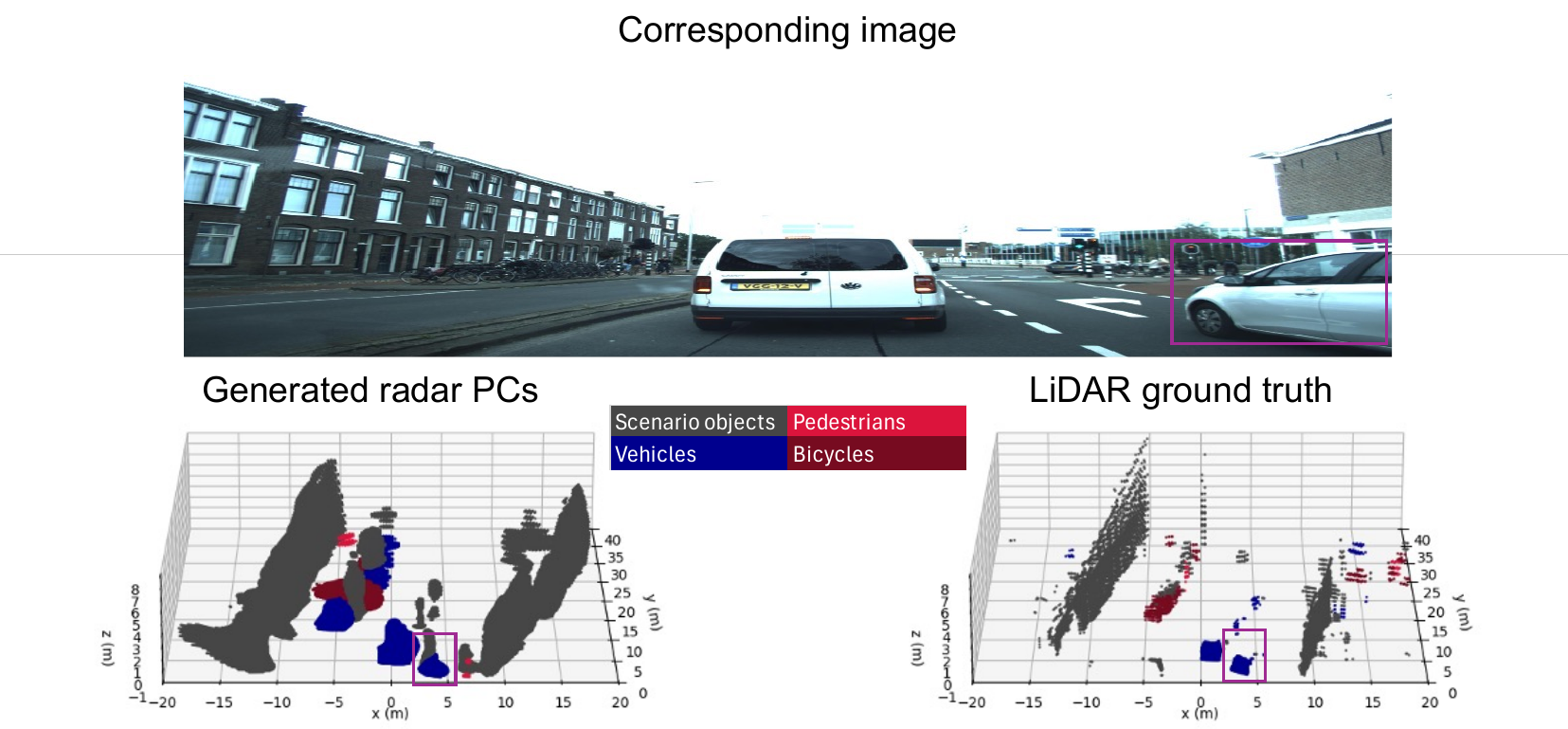}
\caption{Generated radar PCs with class information and LiDAR ground truth are presented for a complex scene. The corresponding RGB image is provided for reference. The color bar with the 4 labeled classes is the same as the one used in the labelling process.}
%The automatic labelling closely aligns with the manual results, with minor discrepancies observed in the green boxes where closely placed bicycles lead to slight variations.}

\label{fig: Visualization results of output}
\end{figure*}

\subsection{Proposed network architecture for semantic segmentation}
The original RAED tensor from the \textit{RaDelft} dataset has dimensions of $N_C \times N_D \times N_A \times N_R$, where the elevation information is embedded within the second dimension of $N_C$, making it less intuitive to use. Moreover, the ground truth generated from LiDAR lacks Doppler information. To address this, the RAED tensor is transformed into a Range-Azimuth-Elevation (RAE) tensor with dimensions $500 \times 240 \times 34$ ($N_R \times N_A \times N_E$). In this transformation, each voxel's value is the average power from the corresponding spatial location in the original radar tensor in polar coordinates.

Starting from this tensor, the proposed semantic segmentation network begins with a 2D backbone to generate two individual feature maps, as shown in Fig. \ref{fig: Block diagram of network structure}. The first branch uses a Feature Pyramidal Network (FPN) \cite{lin2017feature} structure with a Resnet-18 backbone to extract the multi-scale features and finally generate a 3D occupancy latent space with the dimension $N_R\times N_A\times N_E$. The second branch also utilizes an FPN structure with a ResNet-18 backbone, but with different channels to produce the class latent space with dimensions $C\times N_R \times N_A$ ($C=4$ in this work, i.e., $4$ classes of interest). These two feature maps are then combined, utilizing the broadcast property, to construct a 3D semantic information latent space with dimensions $C \times N_R \times N_A \times N_E$.
The second part of the network uses a 3D backbone to integrate the 3D occupancy with class feature maps. A 3D U-Net \cite{cciccek20163d} with five layers is constructed for this purpose. This architecture produces 3D semantic segmentation results with dimensions $C \times N_R \times N_A \times N_E$, and is more computationally efficient than using a 3D convolutional network. 

A combination of weighted cross-entropy (wCE) loss and soft-dice (SDice) loss is used for training. The wCE loss is designed for voxel-wise classification and addresses class imbalance, while the SDice loss is effective for shape segmentation. The network is trained using $80\%$ of the data from five of the seven scenarios, with the remaining $20\%$ reserved for validation. The two remaining scenarios are used for testing as they are completely unseen for the network. %The training was conducted on an NVIDIA A100 GPU provided by the DelftBlue Supercomputer \cite{DHPC2024}, and it took nearly 30 hours to complete 50 epochs.

\subsection{Semantic segmentation results}
\begin{table*}[htbp]
\centering
\caption{Performance of the proposed semantic segmentation network vs variants inspired from the literature. Results in () are for versions of each network trained with the classes `pedestrians' \& `bicycles' combined into a `VRU' class. The best performance for each metric is marked in bold.}
% Two variant models are evaluated to compare with the proposed network. $P_d(All)$, $P_d(Scenario)$, $P_d(Vehicles)$, and $P_d(VRU)$ represent the detection probability of all points, `scenario objects', `vehicles', and VRUs respectively. $P_{fa}(All)$ represents the false alarm for all points. $CD(All)$, $CD(Scenario)$, and $CD(Targets)$ represents the chamfer distance for all points, `scenario objects' and all targets respectively.
\begin{tabular}{lcccccccc}
\toprule
\textbf{Methods} & $P_{d\_\text{All}}\uparrow$ & $P_{fa\_\text{All}}\downarrow$ & $P_{d\_\text{Scenario}}\uparrow$&$P_{d\_\text{Vehicles}}\uparrow$&$P_{d\_\text{VRU}}\uparrow$ &$CD_{\text{All}}\downarrow$ & $CD_{\text{Scenario}}\downarrow$&$CD_{\text{Targets}}\downarrow$ \\ 
\midrule
Baseline (VRU)        & \makecell{63.1\% \\ \textbf{(65.1\%)}} &\makecell{3.55\% \\ (3.09\%)}  &\makecell{60.7\% \\ \textbf{(61.6\%)}} &\makecell{34.8\% \\\textbf{ (47.9\%)}}&\makecell{23.1\% \\ (25.3\%)}&\makecell{2.15m \\ (1.86m)}&\makecell{2.45m \\ (2.15m)}&\makecell{6.73m \\ (5.79m)}\\ 

Baseline + Res (VRU)    & \makecell{63.0\% \\ (63.2\%)}  & \makecell{3.41\% \\ (2.98\%)}& \makecell{60.9\% \\ (58.4\%)} & \makecell{35.9\% \\ (42.6\%)}& \makecell{24.2\% \\ \textbf{(25.8\%)}} & \makecell{1.97m \\ \textbf{(1.77m)}} & \makecell{2.23m \\ \textbf{(2.08m)}} & \makecell{5.95m \\ \textbf{(5.33m)}} \\ 

Variant 1 \cite{roldan2024see} (VRU)      & \makecell{54.2\% \\ (60.4\%)}  & \makecell{3.05\% \\ (3.38\%)}& \makecell{49.9\% \\ (57.9\%)} & \makecell{32.9\% \\ (34.7\%)}& \makecell{21.3\% \\ (22.4\%)} & \makecell{2.34m \\ (2.31m)} & \makecell{2.79m \\ (2.61m)} & \makecell{7.53m \\ (7.40m)} \\ 

Variant 2 \cite{paek2022k} (VRU)      & \makecell{42.4\% \\ (46.8\%)}  & \makecell{\textbf{2.43\%} \\ (2.77\%)}& \makecell{39.1\% \\ (42.1\%)} & \makecell{20.8\% \\ (28.7\%)}& \makecell{13.7\% \\ (16.8\%)} & \makecell{2.81m \\ (2.69 m)} & \makecell{3.55m \\ (3.23 m)} & \makecell{8.14m \\ (7.79m)}  \\ 
\bottomrule
\end{tabular}
\label{tab2}
\end{table*}

\subsubsection{Qualitative results}
The 3D semantic segmentation outputs are transformed into Cartesian coordinates to compare with LiDAR ground truth visually. Fig.~\ref{fig: Visualization results of output} shows the generated radar PCs with class information, the corresponding LiDAR PCs ground truth, and the reference camera image. The results indicate that the generated radar PCs successfully detect most targets within the field of view (FoV). Close vehicles are accurately segmented. Among all vehicles in the FoV, those bounded by the purple box are dynamic, and their segmentation benefits from the radar's Doppler information. More distant pedestrians or vehicles are harder to segment. For static bicycles on the left hand side, some are correctly segmented into their class, while others are incorrectly classified as vehicles due to their dense arrangement. This indicates that radar is more effective at segmenting close-by and dynamic targets compared to others.

\subsubsection{Quantitative results}
Since radar PCs are noisier than those from LiDAR and include reflections also from optically occluded objects, traditional semantic segmentation metrics such as Intersection over Union (IoU) are not suitable. To simultaneously evaluate detection \& segmentation performance, we calculate the detection probability ($P_d$) and false alarm rate ($P_{fa}$) for all classes and for each individual class. Notably, we put the `pedestrians' and `bicycles' in the same class to consider vulnerable road users (VRUs).  Additionally, the Chamfer distance is computed for all classes, `scenario objects' only, and all targets to measure the similarity between their PCs. 

As there are no existing multi-class semantic segmentation methods in 3D space using RAED radar tensors as input and LiDAR PCs as ground truth, direct comparisons are challenging. Thus, we compare our approach with two adapted methods from the literature. The first comparison is with the network proposed by \cite{roldan2024see}, which uses the original data from the \textit{RaDelft} dataset as input and adapts their data-driven detection network into a semantic segmentation network, named `Variant 1'. The second comparison is with \cite{paek2022k}, where the input is made sparse by retaining only the top $10\%$ of values, and the detection head is removed to modify the network into a semantic segmentation network, named `Variant 2'. We also consider an improved baseline, where the 3D U-Net in the 3D backbone is modified with a residual structure, `Baseline + Res'.  For each network, we also train a version where `pedestrians' \& `bicycles' are merged into a `VRU' class.

The performance evaluation uses the following metrics:

\begin{itemize}
    \item $P_{d\_\text{All}}$: Detection probability for all points.
    \item $P_{d\_\text{Scenario}}$: Detection probability for `scenario objects'.
    \item $P_{d\_\text{Vehicles}}$: Detection probability for vehicles.
    \item $P_{d\_\text{VRU}}$: Detection probability for VRUs.
    \item $P_{fa\_\text{All}}$: False alarm rate for all points.
    \item $CD_{\text{All}}$: Chamfer distance for all points.
    \item $CD_{\text{Scenario}}$: Chamfer distance for `scenario objects'.
    \item $CD_{\text{Targets}}$: Chamfer distance for all targets except `scenario objects'.
\end{itemize}

Table \ref{tab2} presents the metrics averaged across all test data. Radar point clouds achieve over 65\% of the LiDAR detection performance and nearly 50\% for $P_{d\_\text{Vehicles}}$. Compared to the two competing methods, the VRU version of the `Baseline + Res' and `Baseline' demonstrate the highest metrics except $P_{fa\_\text{All}}$. The generally low $P_{d\_\text{VRU}}$ comes from the low speed of VRUs, leading to their classification as `scenario objects'. The relatively high value of $CD_{\text{Targets}}$ results from not distinguishing static targets from dynamic ones in the training and test process, causing some static targets to be classified as `scenario objects', i.e., parked cars. The VRU versions generally perform better than the normal ones due to reduced class complexity in the semantic segmentation process.

\section{Conclusions}
An automatic approach for labelling is presented for automotive datasets, using complementary information from camera and LiDAR. The generated labels are used together with related 4D radar data as inputs to a designed semantic segmentation network to associate class labels to each spatial voxel. 
Promising results are shown by applying both approaches to the publicly shared \textit{RaDelft} dataset, with the proposed network achieving more than 65\% in $P_d$, improving 13.2\% in $P_{d\_\text{Vehicles}}$, and reducing 0.54 m in $CD_{\text{All}}$ compared to variants inspired from the literature. The dataset and generated labels
are publicly available at \url{https://data.4tu.nl/datasets/4e277430-e562-4a7a-adfe-30b58d9a5f0a}, with more details of the proposed method in \cite{botao_thesis}.

\newpage
\bibliographystyle{IEEEtran}
\bibliography{reference}

\begin{thebibliography}{10}
\providecommand{\url}[1]{#1}
\csname url@samestyle\endcsname
\providecommand{\newblock}{\relax}
\providecommand{\bibinfo}[2]{#2}
\providecommand{\BIBentrySTDinterwordspacing}{\spaceskip=0pt\relax}
\providecommand{\BIBentryALTinterwordstretchfactor}{4}
\providecommand{\BIBentryALTinterwordspacing}{\spaceskip=\fontdimen2\font plus
\BIBentryALTinterwordstretchfactor\fontdimen3\font minus \fontdimen4\font\relax}
\providecommand{\BIBforeignlanguage}[2]{{%
\expandafter\ifx\csname l@#1\endcsname\relax
\typeout{** WARNING: IEEEtran.bst: No hyphenation pattern has been}%
\typeout{** loaded for the language `#1'. Using the pattern for}%
\typeout{** the default language instead.}%
\else
\language=\csname l@#1\endcsname
\fi
#2}}
\providecommand{\BIBdecl}{\relax}
\BIBdecl

\bibitem{wang2019multi}
Z.~Wang, Y.~Wu, and Q.~Niu, ``Multi-sensor fusion in automated driving: A survey,'' \emph{Ieee Access}, vol.~8, pp. 2847--2868, 2019.

\bibitem{chen2017rethinking}
L.-C. Chen, G.~Papandreou, F.~Schroff, and H.~Adam, ``Rethinking atrous convolution for semantic image segmentation,'' \emph{arXiv preprint arXiv:1706.05587}, 2017.

\bibitem{dosovitskiy2020image}
A.~Dosovitskiy, L.~Beyer, A.~Kolesnikov, D.~Weissenborn, X.~Zhai, T.~Unterthiner, M.~Dehghani, M.~Minderer, G.~Heigold, S.~Gelly \emph{et~al.}, ``An image is worth 16x16 words: Transformers for image recognition at scale,'' \emph{arXiv preprint arXiv:2010.11929}, 2020.

\bibitem{qi2017pointnet++}
C.~R. Qi, L.~Yi, H.~Su, and L.~J. Guibas, ``Pointnet++: Deep hierarchical feature learning on point sets in a metric space,'' \emph{Advances in neural information processing systems}, vol.~30, 2017.

\bibitem{zhao2021point}
H.~Zhao, L.~Jiang, J.~Jia, P.~H. Torr, and V.~Koltun, ``Point transformer,'' in \emph{Proceedings of the IEEE/CVF international conference on computer vision}, 2021, pp. 16\,259--16\,268.

\bibitem{roldan2024deepautomotiveradardetector}
I.~Roldan, A.~Palffy, J.~F.~P. Kooij, D.~M. Gavrila, F.~Fioranelli, and A.~Yarovoy, ``A deep automotive radar detector using the radelft dataset,'' \emph{IEEE Transactions on Radar Systems}, vol.~2, pp. 1062--1075, 2024.

\bibitem{botao_thesis}
\BIBentryALTinterwordspacing
B.~Sun, ``Autolabeling \& semantic segmentation with 4d radar tensors,'' \emph{TU Delft MSc Thesis}, 2024. [Online]. Available: \url{https://repository.tudelft.nl/record/uuid:f01462b1-0446-481e-9333-9b8d3a488f14}
\BIBentrySTDinterwordspacing

\bibitem{shi2020points}
S.~Shi, Z.~Wang, J.~Shi, X.~Wang, and H.~Li, ``From points to parts: 3d object detection from point cloud with part-aware and part-aggregation network,'' \emph{IEEE transactions on pattern analysis and machine intelligence}, vol.~43, no.~8, pp. 2647--2664, 2020.

\bibitem{Geiger2013IJRR}
A.~Geiger, P.~Lenz, C.~Stiller, and R.~Urtasun, ``Vision meets robotics: The kitti dataset,'' \emph{International Journal of Robotics Research (IJRR)}, 2013.

\bibitem{lee2022patchworkpp}
S.~Lee, H.~Lim, and H.~Myung, ``{Patchwork++: Fast and robust ground segmentation solving partial under-segmentation using 3D point cloud},'' in \emph{Proc. IEEE/RSJ Int. Conf. Intell. Robots Syst.}, 2022, pp. 13\,276--13\,283.

\bibitem{jain2023oneformer}
J.~Jain, J.~Li, M.~T. Chiu, A.~Hassani, N.~Orlov, and H.~Shi, ``Oneformer: One transformer to rule universal image segmentation,'' in \emph{Proceedings of the IEEE/CVF Conference on Computer Vision and Pattern Recognition}, 2023, pp. 2989--2998.

\bibitem{domhof2021joint}
J.~Domhof, J.~F. Kooij, and D.~M. Gavrila, ``A joint extrinsic calibration tool for radar, camera and lidar,'' \emph{IEEE Transactions on Intelligent Vehicles}, vol.~6, no.~3, pp. 571--582, 2021.

\bibitem{ester1996density}
M.~Ester, H.-P. Kriegel, J.~Sander, X.~Xu \emph{et~al.}, ``A density-based algorithm for discovering clusters in large spatial databases with noise,'' in \emph{kdd}, vol.~96, no.~34, 1996, pp. 226--231.

\bibitem{lin2017feature}
T.-Y. Lin, P.~Doll{\'a}r, R.~Girshick, K.~He, B.~Hariharan, and S.~Belongie, ``Feature pyramid networks for object detection,'' in \emph{Proceedings of the IEEE conference on computer vision and pattern recognition}, 2017, pp. 2117--2125.

\bibitem{cciccek20163d}
{\"O}.~{\c{C}}i{\c{c}}ek, A.~Abdulkadir, S.~S. Lienkamp, T.~Brox, and O.~Ronneberger, ``3d u-net: learning dense volumetric segmentation from sparse annotation,'' in \emph{Medical Image Computing and Computer-Assisted Intervention--MICCAI 2016: 19th International Conference, Athens, Greece, October 17-21, 2016, Proceedings, Part II 19}.\hskip 1em plus 0.5em minus 0.4em\relax Springer, 2016, pp. 424--432.

\bibitem{roldan2024see}
I.~Roldan, A.~Palffy, J.~F. Kooij, D.~M. Gavrila, F.~Fioranelli, and A.~Yarovoy, ``See further than cfar: a data-driven radar detector trained by lidar,'' in \emph{2024 IEEE Radar Conference (RadarConf24)}.\hskip 1em plus 0.5em minus 0.4em\relax IEEE, 2024, pp. 1--6.

\bibitem{paek2022k}
D.-H. Paek, S.-H. Kong, and K.~T. Wijaya, ``K-radar: 4d radar object detection for autonomous driving in various weather conditions,'' \emph{Advances in Neural Information Processing Systems}, vol.~35, pp. 3819--3829, 2022.

\end{thebibliography}
\end{document}